\documentclass[11pt,a4paper]{article}
\usepackage[hyperref]{emnlp2018}
\usepackage{tempora}
\usepackage[utf8]{inputenc}
\usepackage[english,russian]{babel}
\usepackage{xcolor}
\usepackage{latexsym}
\usepackage{graphicx}
\usepackage{subcaption}
\usepackage{paralist}
\usepackage{siunitx}

\usepackage[normalem]{ulem}
\useunder{\uline}{\ul}{}

\DeclareSymbolFont{extraup}{U}{zavm}{m}{n}
\DeclareMathSymbol{\varheart}{\mathalpha}{extraup}{86}
\DeclareMathSymbol{\vardiamond}{\mathalpha}{extraup}{87}

\usepackage{url}

\newcommand{\ignore}[1]{}

\newcommand{\Sref}[1]{\S\ref{#1}}

\newcommand{\Fref}[1]{Figure~\ref{#1}}

\newcommand{\Tref}[1]{Table~\ref{#1}}

\newcommand{\lexicon}{AgendaLex}

\aclfinalcopy 

\title{Framing and Agenda-setting in Russian News:\\ a Computational Analysis of Intricate Political Strategies}

\author{Anjalie Field$^{\spadesuit}$ ~ Doron Kliger$^{\vardiamond}$ ~Shuly Wintner$^{\vardiamond}$ ~ Jennifer Pan$^{\varheart}$ ~ Dan Jurafsky$^{\varheart}$ ~ Yulia Tsvetkov$^{\spadesuit}$ \\
$^\spadesuit$Carnegie Mellon University ~~ $^\vardiamond$University of Haifa ~~ $^\varheart$Stanford University\\
{ \tt $\{$anjalief, ytsvetko\}@cs.cmu.edu,} \\{\tt \{kliger@econ, shuly@cs\}.haifa.ac.il, \{jp1, jurafsky\}@stanford.edu}}

\date{}

\begin{document}
\maketitle
\begin{abstract}
Amidst growing concern over media manipulation,
NLP attention has focused  on overt strategies like censorship and ``fake news''.
Here, we draw on two concepts from the political science literature to explore subtler strategies for government media manipulation: agenda-setting (selecting what topics to cover) and framing (deciding how topics are covered). We analyze 13 years (100K articles) of the Russian newspaper \textit{Izvestia} and identify a strategy of distraction: articles mention the U.S.\ more frequently in the month directly following an economic downturn in Russia. We introduce embedding-based methods for cross-lingually projecting English frames to Russian, and discover that these articles emphasize U.S.\ moral failings and threats to the U.S. Our work offers new ways to  identify subtle media manipulation strategies at the intersection of agenda-setting and framing.
\end{abstract}

\ignore{Amidst growing concern over media manipulation and disinformation, 
NLP attention has focused primarily on overt strategies like ``fake news'' and censorship.
Here, we draw on two concepts from political science literature to explore
subtler strategies for government media manipulation:
agenda-setting (selecting what topics to cover) and framing (how the topics are covered).
We analyze both of these concepts the Russian newspaper \textit{Izvestia} over a period of 13 years (100K articles). First, we use Granger-causality to establish a casual relationship between how often the news discusses the U.S. and Russia's economic state:
articles mention the U.S. more frequently in the month directly following an economic downturn.
Next, we develop a novel method for projecting framing annotations cross-lingually. We use this method to analyze which frames are salient in U.S.-focused news articles and to what extent the correlation between U.S. news coverage and economic state constitutes a strategy of distraction.
We ultimately combine these analyses and identify subtle media manipulation strategies as the intersection of agenda-setting and framing.}

\section{Introduction}

Authoritarian countries such as Russia and China have received a great deal of attention for trying to control and distort the spread of information through ``fake news'' and censorship. However, authoritarian governments might also use subtle tactics of media manipulation that are much harder to detect, like flooding communication channels with irrelevant information or highlighting particular viewpoints of an event to distract public attention \cite{rozenas2017autocrats, munger2018elites,king2017chinese}. ``Fake news'' can be identified by fact checkers. Censorship can be detected by checking what content is no longer available. However, we have no systematic way of identifying more subtle forms of media manipulation. To date, research has been limited to occasional leaks to reveal ground truth data \cite{king2017chinese}. This paper proposes techniques---grounded in economics and political science---to automatically identify subtle manipulation at scale, and applies these techniques to study Russian media.

These subtle manipulation strategies can be understood through \emph{agenda-setting}---selecting \emph{what} topics to cover---and \emph{framing}---\emph{how} aspects of those topics are highlighted to promote particular interpretations \cite{entman2007framing, ghanem2001convergence}. For example, abortion can be framed in terms of the life of a child or a woman's freedom of choice \cite{tankard2001empirical}. Agenda-setting and framing can have a significant influence on public opinion by attending to particular issues at the exclusion of others \cite{mccombs2002agenda, boydstun2013identifying}. Both concepts have been well-studied in English-speaking democratic countries, but understudied in other settings. Here, we apply these concepts to the study of media manipulation in Russia, particularly as strategies of an autocratic regime.

We focus on Russia, because of intense interest in the way Russia is shaping the global information environment \cite{van2015putin}. Many Russian media outlets are state-owned or heavily influenced by the government. We focus on news coverage from 2003--2016 in one of the most widely-read newspapers in Russia: \textit{Izvestia}. Despite a brief period of autonomy, \textit{Izvestia} has become strongly influenced by the government \cite{jones2002russian}.

Prior work has identified a relationship between negative economic performance in Russia, such as stock market declines, and ``selection attribution'' in state-controlled media outlets, where negative events are blamed on foreign officials while positive events are credited to domestic officials \cite{rozenas2017autocrats}. We build on these findings and investigate the relationship between economic performance, including that of the Russia Trading System Index (RTSI) and gross domestic product (GDP), and news coverage of foreign events. We primarily investigate coverage of the United States because Russia has seen the U.S. as its main rival since the Cold War, and we expect news coverage of foreign events to focus disproportionally on the U.S.

We first establish a strong negative correlation between Russia's economic situation and the proportion of news focused on the U.S.\ (\Sref{sec:agenda_setting}). We then show that the correlation is directed: economic indicators precede (and thereby Granger-cause) the increase in foreign news coverage (\Sref{granger_section}). We consider this a clear case of agenda-setting. We then investigate \textit{how} these news articles \textit{frame} the U.S.
We develop a distant supervision method to  project English framing annotations onto our Russian corpus (\Sref{sec:framing}), and draw on the projected frames to analyze manipulation strategies in news about the U.S. (\Sref{sec:analysis}).

The contributions of this work are manifold:
\begin{inparaitem}[]
\item we show how framing and agenda-setting (concepts traditionally applied to policy debates) can be used to understand media manipulation strategies;
\item we use economic metrics to automate the identification of agenda-setting;
\item we devise a novel method for cross-lingual projection of framing annotations; and
\item we use these annotations to show how agenda-setting is realized.

\end{inparaitem}

\section{Agenda-Setting}
\label{sec:agenda_setting}
All media outlets inevitably use a form of \textit{agenda-setting}: deciding what is ``newsworthy'' by covering some topics at the exclusion of others. Agenda-setting can powerfully sway the focus of public opinion \cite{mccombs2002agenda}. We hypothesize that in countries with weak democratic institutions and in particular, with state-controlled media, the government may actively use agenda-setting to shape public opinion. We observe this phenomenon by comparing how much Russian newspapers describe the U.S.\ and the state of the Russian economy. We then use Granger-causality to show that a decline the Russian stock market is followed by an increase in U.S.\ news coverage.

Our results are  based on a corpus of over 100,000 articles from the newspaper \textit{Izvestia} published in 2003--2016 (see Appendix A for details).

\subsection{Correlations}
\begin{figure}[hbt]
  \vspace*{-8pt}
  \includegraphics[width=0.5\textwidth]{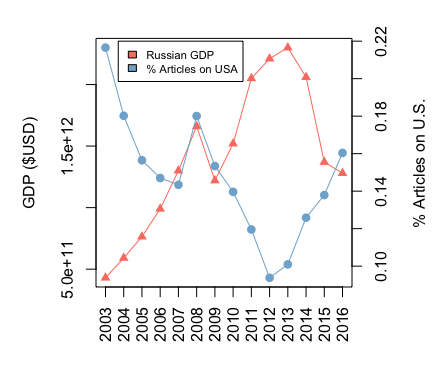}
  \vspace*{-25pt}
  \caption{Proportion of articles that mention the U.S. at least twice (blue) and Russian GDP (red), 2003--2016.}
  \label{fig:yearly}
\end{figure}

We compared the salience of news focused on the U.S.\ with indicators that reflect the economic state of Russia to test our hypothesis: that news coverage of the U.S.\ is used to distract the public from negative economic events.
We first performed an initial, simplistic study of this agenda-setting strategy. 
We define \emph{U.S.\ coverage} as the ratio of \textit{Izvestia} articles that mention the U.S.\  at least twice to the total number of articles in any given time slice (in our initial study, a year). 
We show in \Fref{fig:yearly} the U.S.\ coverage plotted against Russian GDP, in an annual resolution. We find a strong negative Pearson's correlation ($r$=-0.83): mentions of the U.S.\ in \textit{Izvestia} increase as economic indicators deteriorate. The one exception to this trend is 2008, during which there was a high amount of U.S. news coverage and the Russian GDP peaked. This year coincides with both the U.S. financial crisis and the Obama-McCain Presidential election, which would explain a focus on U.S. events regardless of the Russian economic situation. 

\emph{U.S.\ coverage}, measured by counting mentions of the U.S.\ in \textit{Izvestia}, is inversely related to the level of the Russian GDP. This negative correlation indicates the possibility of intentional agenda-setting by the Russian government.
 
We now extend these preliminary results in several ways.
First, we refine the definition of \emph{U.S.\ coverage} by using two metrics:
\begin{inparadesc}
\item[article level,] the number of articles that mention the U.S.\ at least twice normalized by the total number of articles in the time slice; and
\item[word level,] the frequency of the occurrences of the U.S., normalized by the total number of words in the time slice.
\end{inparadesc}
Second, we compare these metrics to \emph{two} economic indicators:  GDP (in USD) and the index of the Russian stock market (RTSI), in rubles.\footnote{Stock market values were obtained from the \href{https://www.moex.com/}{Moscow exchange website}. GDP values were obtained from \href{http://www.oecd-ilibrary.org/}{OECD}.}
Third, we refine the time-resolution and use yearly, quarterly, and monthly time slices.

\Tref{tab:freq_cors} reports the correlations between the two metrics of U.S.\ coverage and  (monthly, quarterly, and yearly) RTSI and GDP values. At all levels, there are strong negative correlations between the proportion of news focused on the U.S.\ and economic state.

\begin{table}[hbt]
\centering
\begin{tabular}{lll}
  \multicolumn{1}{r}{\textbf{Level:}}                      & \textbf{Article} & \textbf{Word} \\\hline
{RTSI (Monthly, rubles)}  & -0.54                  & -0.52               \\
{GDP (Quarterly, USD)} & -0.69                  & -0.65               \\
{GDP (Yearly, USD)}    & -0.83                  & -0.79              
\end{tabular}
\caption{Pearson's correlation between news coverage of the U.S.\ and economic indicators.}
\label{tab:freq_cors}
\end{table}

\subsection{Granger Causality}
\label{granger_section}
Next, we hypothesize that these correlations are in fact directed: a change in the economy is followed by a change in U.S.\ news coverage.
To investigate this hypothesis we employ Granger causality \citep{granger1988some}. The key concept behind Granger causality is that cause precedes effect. Thus, a time series $X$ is said to Granger-cause a times series $Y$ if past values $x_{t-i}$ are significant indicators in predicting $y_t$. First, we computed the article-level ($a_t$) and word-level ($w_t$) metrics at a monthly granularity from 2003 to 2016; we also extracted the RTSI monthly close price (in~USD) for the same time period ($r_t$). We then calculated the percentage change of these series as: $C(w_t) = \frac{w_t}{w_{t-1}} - 1$, and equivalently calculated $C(a_t)$ and $C(r_t)$. By taking the percent change of both series, we control for long term trends (e.g., stock markets tend to trend upwards over time), and instead focus on short-term relations: does a change in the economy directly precede a change in news coverage?

We computed Granger causality between $C(w_t)$ and $C(r_t)$ by fitting a linear regression model:
$$ C(w_t) = \sum_{i=1}^m\alpha_i(C(w_{t-i})) + \sum_{j=1}^n\beta_j(C(r_{t-j})) $$
where $m$ and $n$ denote how far back in time we look (denoted as $m$-lag or $n$-lag). We can say that $r_t$ Granger-causes $w_t$ if we find that $\beta$ is significantly different from zero.

Tables~\ref{granger_word} and~\ref{granger_article} report the results. A $p$-value $\le 0.05$ indicates significance; thus we find 1-lag RTSI values Granger-cause coverage of U.S. news by both the word-level and article-level metrics. Importantly, the $r_{t-1}$ coefficient is negative, which indicates that a decline in the stock market is followed by an increase in U.S. news coverage. In the 2-lag analysis, the $r_{t-2}$ values are not significant, which suggests that the changes in news coverage follow changes in the stock market within one month.\footnote{We  computed Granger causality at a quarterly and yearly level and found no significant causal relationship. This result is unsurprising; the monthly analysis suggests trends in news coverage are largely driven by the previous month, so we would not expect  causality at a quarterly or yearly level.} For completeness, we also computed Granger causality in the reverse direction: i.e., does a change in U.S. news coverage Granger-cause a change in the stock market? As expected, we found no significant results.

\begin{table}[hbt]
\centering
\begin{tabular}{lllll}
                                           & \multicolumn{2}{c}{\textbf{1-Lag}}                                                & \multicolumn{2}{c}{\textbf{2-Lag}}                                                \\ 
\multicolumn{1}{l}{}                     & \multicolumn{1}{c}{$\alpha$; $\beta$} & \multicolumn{1}{l}{\textbf{p-Value}} & \multicolumn{1}{c}{$\alpha$; $\beta$} & \multicolumn{1}{l}{\textbf{p-Value}} \\ \hline
\multicolumn{1}{l}{\textbf{$w_{t-1}$}}  & \multicolumn{1}{l}{-0.233}             & \multicolumn{1}{l}{0.003}          & \multicolumn{1}{l}{-0.320}             & \multicolumn{1}{l}{0.00005}           \\ 
\multicolumn{1}{l}{\textbf{$w_{t-2}$}}  & \multicolumn{1}{l}{-}                    & \multicolumn{1}{l}{-}                & \multicolumn{1}{l}{-0.301}             & \multicolumn{1}{l}{0.0001}           \\
\multicolumn{1}{l}{\textbf{$r_{t-1}$}} & \multicolumn{1}{l}{-0.352}             & \multicolumn{1}{l}{0.0334}          & \multicolumn{1}{l}{-0.369}             & \multicolumn{1}{l}{0.024}           \\
\multicolumn{1}{l}{\textbf{$r_{t-2}$}} & \multicolumn{1}{l}{-}                    & \multicolumn{1}{l}{-}                & \multicolumn{1}{l}{-0.122}             & \multicolumn{1}{l}{0.458}           \\
\end{tabular}
\caption{Granger causality between \% change in RTSI and frequency of USA (word level).}
\label{granger_word}
\end{table}

\begin{table}[hbt]
\centering
\begin{tabular}{lllll}
                                           & \multicolumn{2}{c}{\textbf{1-Lag}}                                                & \multicolumn{2}{c}{\textbf{2-Lag}}                                                \\ 
\multicolumn{1}{l}{}                     & \multicolumn{1}{c}{$\alpha$; $\beta$} & \multicolumn{1}{l}{\textbf{p-Value}} & \multicolumn{1}{c}{$\alpha$; $\beta$} & \multicolumn{1}{l}{\textbf{p-Value}} \\ \hline
\multicolumn{1}{l}{\textbf{$a_{t-1}$}}  & \multicolumn{1}{l}{-0.222}             & \multicolumn{1}{l}{0.005}          & \multicolumn{1}{l}{-0.290}             & \multicolumn{1}{l}{0.000289}           \\
\multicolumn{1}{l}{\textbf{$a_{t-2}$}}  & \multicolumn{1}{l}{-}                    & \multicolumn{1}{l}{-}                & \multicolumn{1}{l}{-0.270}             & \multicolumn{1}{l}{0.000634}           \\
\multicolumn{1}{l}{\textbf{$r_{t-1}$}} & \multicolumn{1}{l}{-0.311}             & \multicolumn{1}{l}{0.035}          & \multicolumn{1}{l}{-0.329}             & \multicolumn{1}{l}{0.0267}           \\
\multicolumn{1}{l}{\textbf{$r_{t-2}$}} & \multicolumn{1}{l}{-}                    & \multicolumn{1}{l}{-}                & \multicolumn{1}{l}{-0.091}             & \multicolumn{1}{l}{0.543}           \\
\end{tabular}
\caption{Granger causality between \% change in RTSI and frequency of USA (article level).}
\label{granger_article}
\end{table}

\section{Framing Analysis}
\label{sec:framing}
We  hypothesize that \textit{framing} can further our understanding of why Russian media focuses on the U.S.\ during economic downturns. By identifying
 common frames in news coverage of the U.S., we  see
how the concepts of agenda-setting and framing work together to manipulate public attention. In this section, we first define the concept of framing and demonstrate why existing methods are insufficient for analysis of the \textit{Izvestia} corpus.  
We then present a new method for analyzing frames and evaluate it quantitatively through hand-annotations and qualitatively through a series of examples. Finally, we use this method to contextualize strategies of media manipulation in the \textit{Izvestia} corpus.

\subsection{Background on Framing Analyses}
\label{sec:framing-background}

While agenda-setting broadly refers to what topics a text covers, framing refers to which \textit{attributes} of those topics are highlighted. Several aspects of framing make the concept difficult to analyze. First, just defining framing has been ``notoriously slippery'' \cite{boydstun2013identifying}. Frames can occur as stock phrases, i.e. ``death tax'' vs. ``estate tax'', but they can also occur as broader associations or sub-topics \cite{tsur2015frame, mccombs2002agenda}. Frames also need to be distinguished from similar concepts, like sentiment and stance. For example, the same frame can be used to take different stances on an issue: one politician might argue that immigrants boost the economy by starting new companies that create jobs, while another might argue that immigrants hurt the economy by taking jobs away from U.S. citizens \cite{baumer2015testing, gamson1989media}. Finally, unlike classification tasks where each article is assigned to a single category, most articles employ a variety of frames \cite{ghanem2001convergence}.

Recent work has attempted to address these conceptual challenges by defining broad framing categories. The Policy Frames Codebook defines a set of 15 frames (one of which is ``Other'') commonly used in media for a broad range of issues \cite{boydstun2013identifying}. In a follow-up work, the authors use these frames to build The Media Frames Corpus (MFC), which consists of articles related to 3 issues: immigration, tobacco, and same-sex marriage \cite{card2015media}. About 11,900 articles are hand-annotated with frames: annotators highlight spans of text related to each frame in the codebook and assign a single ``primary frame'' to each document. However, the MFC, like other prior framing analyses, relies heavily on labor-intensive manual annotations.

The primary automated methods have relied on probabilistic topic models \citep{tsur2015frame,boydstun2013identifying,nguyen2013lexical, roberts2013structural}. Although topic models can show researchers what themes are salient in a corpus, they have two main drawbacks: they tend to be corpus-specific and hard to interpret. Topics discovered in one corpus are likely not relevant to a different corpus, and it is difficult to compare the outputs of topic models run on different corpora.  Other automated framing analyses have used the annotations of the Media Frame Corpus to predict the primary frame of articles \cite{card2016analyzing, ji2017neural}, or used classifiers to identify language specifically related to framing \cite{baumer2015testing}. Importantly, all of these methods focus exclusively on English data sets. While unsupervised methods like topic models can be applied to other languages, any supervised method requires annotated data, which does not exist in other languages.

\subsection{Framing Analysis Methodology}

Our goal is to develop a method that is easy to interpret and applicable across-languages. In order to ensure our analysis is interpretable, we ground our method using the annotations of the Media Frames Corpus. However, because the MFC is entirely in English and our test corpora is in Russian, we cannot use a fully supervised method. Instead, we use the MFC annotations to derive lexicons for each frame, which we then translate into Russian. We use query-expansion to reduce the noisiness of machine translation and make the lexicons specific to the \textit{Izvestia} corpus, rather than specific to the MFC. We evaluate the derived lexicons in English and in Russian. Finally, we use these lexicons to analyze frames in \textit{Izvestia} and identify strategies of media manipulation. Our method allows for in-depth analysis by identifying primary and secondary frames in a document and specific words that signify frames.

\paragraph{Generating framing lexicons}
Although our primary test corpus is in Russian, we also use English test corpora for evaluation; thus, we describe our method as applicable to either language. First, we use the MFC annotations to derive a lexicon of English words related to each frame in the Policy Frames Codebook.
For a given frame $F$ we measure pointwise mutual information \citep{church1990word} for each word in the corpus as:
\[
I(F,w) = log\frac{P(F,w)}{P(F)P(w)} = log\frac{P(w \mid F)}{P(w)} 
\]
We estimate $P(w|F)$ by taking all text segments annotated with frame $F$, and computing $\frac{Count(w)}{Count(allwords)}$. We similarly compute ${P(w)}$ from the entire corpus. We then use the 250 words with the highest $I(w,F)$ as the base framing lexicon for frame $F$, denoted $F_{base}$. We discard all words that occur in fewer than 0.5\% of documents or in more than 98\% of documents.

\paragraph{Translation and extension of framing lexicons}
Next, we use query-expansion to alter $F_{base}$, with the goal of generalizing the lexicon.
Without this step, our lexicons are biased towards words common in English news articles, particularly words specific to the 3 policy issues in the MFC.

When our test corpus is in a different language (i.e. Russian), we use Google Translate to translate $F_{base}$ into the new language. We restrict our vocabulary to the 50,000 most frequent words in the test corpus. 

Then, to perform the query-expansion, we train 200-dimensional word embeddings on a large background corpus in the test language, using  CBOW with a 5-word context window \cite{mikolov2013efficient}. We compute the center of each lexicon, $c$, by summing the embeddings for all words in the lexicon. We then identify up to the $K$ nearest neighbors to this center, determined by the cosine distance from $c$, as long as the cosine distance is not greater than a manually-chosen threshold ($t$).\footnote{When the test corpus is in English, we set $t$ to 0.4 and $K$ to 500 and we add the identified neighbors to $F_{base}$. When our test corpus is in Russian, we choose to discard our base lexicon, to prevent the final lexicons from being too U.S.-specific. Instead, we set $t$ to 0.3 and $K$ to 1000, which increases the number of neighbors identified, and we keep only these neighbors in the final lexicon.}
We again filtered the final set by removing all words that occur in fewer than 0.5\% of documents or in more than 98\% of documents.

The final lexicons contain between 100 and 300 words per frame. 
Table~\ref{tab:example_lex} depicts a few examples of lexicon words extracted from the MFC, and words in our final lexicons. We can observe that words in $F_{rus}$ are closely related to words in $F_{base}$, but also specific to Russian culture and politics.

We consider a document to employ a frame $F$ if the document contains at least 3 instances of a word from $F$'s lexicon. We assign the primary frame of a document to be its most common frame, determined by the number of words from each framing lexicon in the document .\footnote{We do not generate a lexicon for the ``Other'' frame, and instead assign a document's primary frame as ``Other'' only if it does not contain at least 3 words from any framing lexicon. Throughout this process, we use small subsets of the ``tobacco'' articles for parameter tuning.}


\begin{table}[hbt]
\centering

\begin{small}

\begin{tabular}{l@{\hspace{40pt}}l}
  $F_{base}$       & $F_{rus}$    \\ \hline
          \multicolumn{2}{c}{\textbf{Political}} \\
                 republican-controlled    &  bills  \\
                 filibuster    &  conservative   \\
                 gubernatorial &  parlimentary \\
          \multicolumn{2}{c}{\textbf{Economic}} \\
                 cents    &  deductions  \\
                 holdings     &  tax \\  
                 profitable   &  fines  \\  
           \multicolumn{2}{c}{\textbf{Public Sentiment}} \\
                 gallup        &  activism \\
                 demonstrators       &  facsim \\
                 rallied       &  vote \\      
\end{tabular}

\end{small}

\caption{Example lexicons extracted from the MFC and transfered to the \textit{Izvestia} corpus.}
\label{tab:example_lex}
\end{table}

\section{Evaluation of Framing Lexicons}
\label{sec:lex-evaluation}
We can evaluate the English lexicons using annotated data from the MFC. For the Russian lexicons, since we do not have annotated Russian data, we instead conduct an annotation task. These evaluation metrics determine how well our method captures which frames are present in a text. Finally, we also qualitatively compare our method to existing methods for framing analysis, specifically topic models.

\paragraph{English Evaluations}
We first evaluate our lexicons on two tasks using the MFC annotations: primary frame identification and identification of all frames in a document.

Primary frame identification is  a 15-class classification problem. Two prior studies evaluate models on this task: \citet{card2016analyzing} and \citet{ji2017neural}. Following these studies, we evaluate our model using 10-fold cross-validation on only the ``Immigration'' subset of the MFC. We use 9 folds to generate framing lexicons and the 10th fold to evaluate. To train word embeddings, we use the entire MFC corpus combined with over 1 million New York Times articles from 1986 - 2016 \citep{fast2017long}. Table \ref{tab:primary} shows the accuracy of our model. Our results outperform \citet{card2016analyzing} and are comparable to \citet{ji2017neural}. Furthermore, unlike prior methods, our method is able to transfer to different domains and languages without needing further annotated data. 

\begin{table}[hbt]
\centering
\begin{tabular}{lr}
\ \citet{ji2017neural}       & 58.4  \\
\ \citet{card2016analyzing}  & 56.8 \\
\ Our model                  & 57.3    \\
\end{tabular}
\caption{Accuracy of primary frame classification.}
\label{tab:primary}
\end{table}

However, our main interest is in measuring the salience of frames in general, not merely focusing on the primary frame. Thus, we also use our lexicons to identify the presence of any frames in a document. As the MFC has multiple annotators, we define a frame to be present in a document if any annotator identified the frame, and use this as gold standard test data. In evaluating our lexicons, we consider a frame to be present in a document if the document contains at least 3 tokens from the frame's lexicon.

To the best of our knowledge, identifying all frames in a document is a new task that was not attempted in prior work. Thus, we use a logistic regression model with bag-of-word features as a standard baseline. As above, we evaluate using 10-fold cross validation on the ``Immigration'' subset of the MFC. Table \ref{tab:all_frames} shows that our method outperforms the baseline, with the exception of 2 frames, even though the baseline is fully supervised, whereas our method is distantly supervised. We note that the poorest performing frames, ``External Regulation and Reputation'' and ``Morality'' are the frames which are least common in this subset of the data -- each frame occurs in fewer than 500 articles. When we run the same 10-fold cross validation evaluation on the ``Samesex'' subsection of the MFC, where the ``Morality'' frame occurs in over 1000 articles, we achieve a higher F1 score (0.65).

\begin{table}[hbt]
\centering
\begin{tabular}{lll}

	& \textbf{Ours}	&\textbf{Baseline}	\\
Capacity \& Resources	&\textbf{0.53}	&0.48 \\
Crime \& Punishment	&\textbf{0.78}	&0.76 \\
Cultural Identity	    &0.57	&\textbf{0.62} \\
Economic	            &\textbf{0.69}	&0.67 \\
External Regulation	    &0.25	&\textbf{0.47} \\
Fairness \& Equality	&\textbf{0.50}	&0.44 \\
Health \& Safety	    &\textbf{0.58}	&0.53 \\
Legality \& Constitutionality &\textbf{0.80}	&0.76 \\
Morality	            &\textbf{0.31}	&0.25 \\
Policy Prescription  	&\textbf{0.72}	&0.69 \\
Political	            &\textbf{0.80}	&0.77 \\
Public Sentiment	    &\textbf{0.54}	&0.47 \\
Quality of Life	        &\textbf{0.65}	&0.63 \\
Security \& Defense	&0.63	&0.63 \\

\end{tabular}
\caption{F1 Scores for identification of all frames in a document.}
\label{tab:all_frames}
\end{table}

\paragraph{Russian Evaluations}
Next, we evaluate the quality of our method on the Russian data set. Unlike in English, we do not have frame-annotated data in Russian. We instead performed the intruder detection task, an established method for evaluating topic models \citep{chang2009reading}. For each frame $F$ we randomly sampled 5 words from the framing lexicon $F_{rus}$ and 1 word from the lexicon of a different frame, which has no overlap with $F_{rus}$. We then presented two (native Russian speaking) annotators with the frame heading and the set of 6 words, and asked them to choose which word did not belong in the set. We evaluated 15 sets or 75 words per frame.

Framing can be subjective, and we do not necessarily expect annotators to interpret frames in the same way. 
We calculate two forms of accuracy: ``soft'', whether \emph{any} annotator correctly identified the intruder; and ``hard'', whether both did. We also report average precision as defined in \cite{chang2009reading}, i.e. the average number of annotators that correctly identified the intruder, averaged across all sets.

We briefly summarize results here and report them fully in Appendix B. All accuracies are significantly better than random guessing, and no soft accuracy falls below 60\%. Only two frames have an average accuracy $\le60\%$, ``Fairness and Equality'' and ``Morality'', both very abstract concepts. In these frames, we also see a larger difference between hard and soft accuracies, which reflects the subjectivity of framing. The MFC annotators sometimes disagreed on the correct annotations, even after discussing their disagreements \citep{boydstun2013identifying}. Thus, we can attribute some of the differences between hard and soft accuracies to this subjectivity.

\paragraph{Qualitative Comparison to Structured Topic Models}
We also qualitatively compared the information our framing lexicons provide with information provided by a Structured Topic Model (STM) \cite{roberts2013structural}. We find that our approach is better able to capture frames the way a reader might conceptualize them, whereas topic models are useful for finding fine-grained corpus-specific topics.

Topic models are  a common way to analyze frames in a text \cite{nguyen2013lexical}; the STM specifically allows correlation between topics and covariates. We trained an STM with 10, 15, 20, 25, and 50 topics on U.S.-focused articles in the \textit{Izvestia} corpus, including publication date (month and year) as a covariate. We selected the 20 topic model as having the most coherent topics. Throughout this section, we refer to topics using their most representative words as determined by the ``Lift'' metric \cite{roberts2013structural}.


We randomly selected a sample document for each primary frame
to investigate. The framing lexicons are able to connect corpus-specific vocabulary to higher-level concepts. For example, an article describing movies about the U.S. prison facility at Guantanamo Bay has two main STM topics: [laden, sentence, prison] and  [author, viewer, filming]. Similarly, the framing lexicons identify `Cultural Identity'' as the primary frame. However, a secondary frame in the document is ``Morality'', captured by words: writer, form, Christ, art. 
While both the STM and the framing lexicons capture major details of the article, the framing lexicons additionally  tie the article to morality,  because words like ``art'' in this corpus are often signs of a moral framework. 

Nevertheless, when the STM identifies a topic similar to a frame, we find correlations with the related lexicon, i.e. there is a 0.75 correlation between the frequency of words in the Legality, Constitutionality, Jurisdiction lexicon and the monthly average proportion of each document assigned to the topic [yukos, bill, legislation].

Additionally, the framing lexicons tend to have higher precision in identifying relevant articles than the STM. Topics are commonly identified by their most probably words, which may not occur at all in documents associated with the topic. For example, the STM assigns an article about smoking policies in the U.S. to 3 main topics:  [laden, sentence, prison], [kosovo, falcons, because], [author, viewer, filming], none of which are closely related to the article. In contrast, because assignments to the framing lexicons are made directly from words in the lexicon, we can be confident that articles assigned to each frame have words from the actual lexicon, and are very likely related to the frame. The framing lexicons assign the primary frame as ``Policy'' for this article, which is a good fit. Neither method captures that the article is also related to health.

Finally, the STM is useful for finding fine-grained topics, beyond the Policy Frames Codebook. For example, we find a ``sports'' topic: [match, nhl, team]. These topics tend to be corpus-specific and more concrete than the framing lexicons: no STM topic captures ``Quality of Life''.

\section{Identifying Media Manipulation}
\label{sec:analysis}
We first use the generated framing lexicons to determine which frames are frequently associated with the U.S. We then break the frames into finer-categories and manually look at sample articles to determine why associating these frames with the U.S. constitutes a media manipulation strategy. We find that as the stock market declines, not only is news focused more on the U.S., but also emphasizes threats to the U.S.

\subsection{Salient frames}

To estimate which frames are associated with the U.S. we compute \emph{normalized pointwise-mutual information} (nPMI) between the U.S. and each frame $F$\footnote{As above, we consider an article to be U.S.-focused if it mentions the U.S. at least 2 times, and we consider an article to employ frame $F$ if it uses at least 3 words from $F$'s lexicon.} 
by mapping the mutual information score onto a [-1,1] scale. A value of 1 represents complete co-occurrence; a value of 0 represents complete independence. By using nPMI, we measure which frames are \textit{overrepresented} in U.S.-focused news, as compared to other news.

\begin{figure}[hbt]
\centering
  \includegraphics[width=0.8\columnwidth]{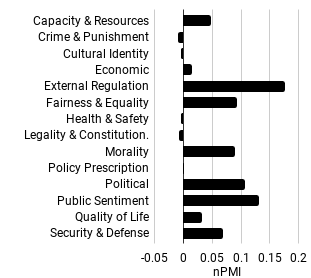}
  \caption{nPMI between U.S. and each frame.}
  \label{fig:pmi_all}
\end{figure}

Figure~\ref{fig:pmi_all} shows the nPMI score between the U.S. and each frame for all articles in our corpus. As any news article about the U.S. is by definition externally focused,   the frame with the strongest association is unsurprisingly ``External Regulation and Reputation''. Other frames with strong associations include ``Morality'', ``Political'', ``Public Sentiment'', and ``Security and Defense''. These frames demonstrate what type of news events in the U.S are reported in Russia. As an example, we look at an article that uses a combination of these frames. The article describes cooling relations between Russia and the U.S. It explains that anti-Russian sentiment will be prevalent in the U.S. during upcoming elections, when politicians on both sides will play the ``Russia card''. It ultimately attributes the cooling relations to a mismatch of values and ideology between the two nations. The framing lexicons well-capture the numerous themes in this article. Specifically, the frames identified and the related framing lexicon words are:

\textit{Political}: electorate, election, former, pre-election, political scientists, congress, president, post, bush

\textit{Public Sentiment}: elections, campaign, pre-election, democrats, republicans


\textit{External Regulation and Reputation}: west, war, former, washington, politics, summit, exacerbation, west, decision, bush, president

\textit{Morality}: peace, sins, ideals, love, values

\textit{Fairness and Equality}:  politics, love, values

This article uses several strategies to promote unity in Russia and actively separate Russia from Western culture, including criticizing American politics and emphasizing a difference of values. Russian articles use a combination of frames to describe the U.S., which demonstrates the importance of looking at all frames in a document, rather than just the primary frame. In the following sections, we provide additional examples demonstrating how combinations of frames can generate anti-U.S. sentiment.

\subsection{Salient words within frames}

We expect different aspects of frames to be foregrounded
during economic upturns than in downturns. To investigate these differences, we  define a set of months $M_t^+$, as the 10\% of months where RTSI showed the greatest growth, and a corresponding set $M_t^-$ where RTSI showed the greatest decline. We then take $M_{t+1}^+$ as the month directly following every month in $M_t^+$, and we similarly define $M_{t+1}^-$.
From the analysis in \Sref{granger_section}, we expect media manipulation strategies to decline from $M_t^+$ to $M_{t+1}^+$, and increase from $M_t^-$ to $M_{t+1}^-$.
For each frame, we take the subset of U.S.-focused articles that use the frame. Then, we use log odds with a Dirichlet prior \citep{jurafsky2014,monroe_colaresi_quinn} to identify words that are overrepresented or underrepresented from $M_t^+$ to $M_{t+1}^+$ and from $M_t^-$ to $M_{t+1}^-$.\footnote{We take the 500 words with the largest increase in salience from $M_t^-$ to $M_{t+1}^-$ intersected with the 500 words with the largest decrease in salience from $M_t^+$ to $M_{t+1}^+$.} Thus, for each frame, we identify words which become more common after a stock market downturn and become less common after a stock market upturn. We refer to these words as \lexicon.

We found the Security and Defense \lexicon\ and the Crime and Punishment \lexicon\ to be surprisingly coherent, both containing words related to terrorism and countries enemy to the U.S., including bombs, missiles, Guantanamo, North Korea, Iraq, etc. We found a correlation of -0.49 between the frequency of words from the Security and Defense \lexicon in U.S.-focused articles and the RSTI (-0.49). A 1-lag Granger causality test (to what extent does a change in RTSI Granger-cause a change in the prevalence of the Security and Defense \lexicon?) has a p-value of 0.0051. As the stock market declines, not only does the news focus more on the U.S., the news focuses specifically on terrorists and other enemies to the U.S. In the next section, we refine this conclusion by looking at sample articles.

\subsection{Examples of framing during downturns}

By reading sample articles from months just after stock market downturns that used words from  the Security and Defense lexicon and \lexicon, we identified three common strategies for distracting Russian citizens from negative economic events: villainizing the U.S., describing threats to the U.S., and promoting the Russian military over the U.S. military.

First, some articles focus on immoral actions of the U.S. military, describing U.S. troops as ``Nazi'', or U.S. campaigns in Iraq as ``barbaric'' or causing ``horror and outrage throughout the world". 
Others discuss  Guantanamo Bay,  employing the ``Morality'' or ``Legality, Constitutionality, Jurisdiction'' frames. 
By portraying the U.S. government negatively,  actions of the Russian government  appear positively by comparison. Promoting unity by presenting an external enemy is a well-studied political strategy.

Second, many articles, often in passing, described threats to the U.S. An article about terror attacks in Paris  mentions U.S. involvement with words from the Crime and Punishment and Security and Defense lexicons: terrorist, terrorists, special services. An article about the conflict between Israel and Palestine describes increased security in the U.S.  An article about a  U.S. military operation refers more directly  to threats to the U.S. by claiming the killed terrorist will simply be replaced  ``and everything will start afresh - explosions, chases, roundups...unlucky businessmen, successful terrorists''. The articles  portray the U.S. as an unsafe place to live, making Russia seem like a preferable home.

A third type of article also presents Russia as safe by downplaying  U.S. military threat: ``the missile defense system of the USA does not pose a real threat to Russia's strategic nuclear forces.'' or describing the growth of Russian technology compared to `impotent' American counterparts.


\section{Related Work}
Most studies on Russian media manipulation focus on state-owned television networks, such as \textit{Channel 1} and \textit{RT}. Strategies identified in these outlets include spreading confusion \citep{paul2016russian} and ``selection attribution'', in which negative economic events are attributed to foreign entities and positive events are attributed to Russian officials \citep{rozenas2017autocrats}. 
Similar strategies have been identified in social media in China and in Venezuela: these regimes flood communication channels with irrelevant information or general ``cheerleading'', presumably to distract the public from current events \citep{king2017chinese,munger2018elites}. We expand on these analyses as we study manipulation strategies in a more automated way, through Granger-causality and framing lexicons. We further draw parallels between these strategies and theories of agenda-setting and framing.

Furthermore, our method for analyzing frames contributes to the growing body of work on automated-framing analysis \citep{nguyen2013lexical, boydstun2013identifying, card2016analyzing, baumer2015testing}. While past work uses fully-supervised methods, which are not applicable to languages lacking training data, or unsupervised topic models, which can be difficult to interpret, we take a semi-supervised approach: using statistical metrics and word embeddings to generate corpus-specific lexicons based on common frameworks.
%

We additionally integrate our framing and agenda-setting analysis with economic indicators through the concept of Granger causality. While the concept of Granger causality is not new in economics, 
it is less common in NLP and social sciences. Moreover, modeling relationships between news and economic indicators is a relatively recent area. Most research has focused on using text to predict economic indicators using a variety of features, from frequencies of keywords to sentiment of social media posts \citep{nardo2016walking}. \citet{kang2017detecting} combine text and Granger causality for a different task: automatically explaining causes of time series events. Our study differs from past work in that we reverse the direction: rather than using news articles to model changes in economic data, we use economic data to show changes in news articles.

\section{Conclusions}

We show that natural language technology, in addition to its ability to address overt manipulation strategies like ``fake news'' and censorship,  has the potential to shed light on  more subtle political manipulation strategies, specifically distraction. We  offer a way to define these strategies by drawing on  social science theories of agenda-setting and framing, combining them with a novel methodology for cross-lingual projection of framing annotations. 
We investigate how the resulting frames are used in the Russian newspaper \textit{Izvestia}, and show that it reports on negative events in the U.S.\ as a way of distracting from economic downturns in the Russian economy. 

Our approach and our findings serve as a starting point for further research on automating the identification and analysis of media manipulation strategies. 
These include identification of more nuanced framing strategies, such as mitigation, projections of power among entities mentioned in news, identification of over- and under-represented events, and in general, detection of biases in news articles as a means for understanding trustworthiness in media reports.

\section*{Acknowledgments}


We gratefully acknowledge our helpful reviewers and annotators and Ethan Fast for providing the NYT corpus. This research was supported by Grant No.~2017699 from the United States-Israel Binational Science Foundation (BSF), by Grant No.~IIS1812327 from the United States National Science Foundation (NSF), and by the Stanford Cyber Initiative. Further, this material is based upon work supported by the NSF Graduate Research Fellowship Program under Grant No. DGE1745016. Any opinions, findings, and conclusions or recommendations expressed in this material are those of the authors and do not necessarily reflect the views of the NSF.


\bibliography{emnlp2018}
\bibliographystyle{acl_natbib_nourl}
\clearpage
\section*{Appendix A}

\begin{table}[h]
\centering
\begin{tabular}{lr}
\# Types                      &  1,013,024  \\
\# Tokens                      & 87,761,626 \\
\# Articles                   & 118,532    \\
Average \# Articles per month & 718      
\end{tabular}
\caption*{Overview of Izvestia corpus}
\end{table}

\paragraph{Preprocessing}
We identified named entities with \href{https://github.com/ispras/texterra-py}{the ISPRAS (texterra) API} and then
 manually grouped country mentions i.e., collapsing ``U.S.A.'', and ``Americans'' to a single label. We paid particular attention to words referring to the U.S. or Russia and allowed less clean references to other countries. Finally, we tokenized and lowercased all text.

\section*{Appendix B}

\begin{table}[hbt]
\centering
\begin{tabular}{llll}
                             & \textbf{Hard} & \textbf{Soft} & \textbf{Avg.}  \\
{Capacity \& Resources}      & 60.00                  & 93.33         & 76.67 \\
{Crime \& Punishment}        & 93.33                  & 93.33         & 93.33 \\
{Cultural Identity}          & 73.33                  & 100           & 86.67 \\
{Economic}                   & 100                    & 100           & 100 \\
{External Regulation}        & 93.33                  & 100           & 96.67 \\
{Fairness \& Equality}       & 20.00                  & 60.00         & 40.00 \\
{Health \& Safety}           & 93.33                  & 93.33         & 93.33 \\
{Legality \& Constitution.}  & 86.67                  & 93.33         & 90.00 \\
{Morality}                   & 46.67                  & 73.33         & 60.00 \\
{Policy Prescription}        & 86.67                  & 100           & 93.33 \\
{Political}                  & 73.33                  & 86.67         & 80.00 \\
{Public Sentiment}           & 53.33                  & 86.67         & 70.00 \\
{Quality of Life}            & 66.67                  & 93.33         & 80.00 \\
{Security \& Defense}        & 60.00                  & 66.67         & 63.33 \\
\end{tabular}
\caption*{Soft and hard accuracy scores (\%) and average precision in Russian framing lexicon intruder detection task}
\label{precision_recall}
\end{table}

\end{document}